\documentclass[sigconf]{acmart}

\usepackage{booktabs} 
\usepackage[colorinlistoftodos]{todonotes}
\usepackage{algorithmic}
\usepackage[linesnumbered, ruled, vlined,boxed]{algorithm2e}
\usepackage{multirow}
\usepackage{tablefootnote}
\usepackage{graphicx}
\usepackage{subfigure}
\usepackage{flushend}
\usepackage{color}
\usepackage{enumitem}
\usepackage{array, makecell}
\usepackage{footmisc}

\usepackage{xcolor}
\usepackage{xspace}
\usepackage{arydshln}

\newcommand{\xz}[1]{\textbf{\color{blue}[(Xu: #1 )]}}

\usepackage[]{threeparttable}

\newcommand{\hide}[1]{} 
\newcommand{\vpara}[1]{\vspace{0.07in}\noindent\textbf{#1 }}
\newcommand{\beq}[1]{\begin{equation}#1\end{equation}}
\newcommand{\model}{inverse prompting\xspace}
\newcommand{\modelst}{Inverse prompting\xspace}

\copyrightyear{2021}
\acmYear{2021}
\setcopyright{acmlicensed}\acmConference[KDD '21]{Proceedings of the 27th ACM SIGKDD Conference on Knowledge Discovery and Data Mining}{August 14--18, 2021}{Virtual Event, Singapore}
\acmBooktitle{Proceedings of the 27th ACM SIGKDD Conference on Knowledge Discovery and Data Mining (KDD '21), August 14--18, 2021, Virtual Event, Singapore}
\acmPrice{15.00}
\acmDOI{}
\acmISBN{}

\setcopyright{acmcopyright}
\copyrightyear{2021}
\acmYear{2021}

\begin{document}

\settopmatter{printacmref=false}
\setcopyright{none}
\renewcommand\footnotetextcopyrightpermission[1]{}
\pagestyle{plain}

\newcommand\blfootnote[1]{%
  \begingroup
  \renewcommand\thefootnote{}\footnote{#1}%
  \addtocounter{footnote}{-1}%
  \endgroup
}

\fancyhead{}

\title{Controllable Generation from Pre-trained Language Models via Inverse Prompting}

\author[ X. Zou, D. Yin, Q. Zhong, M. Ding, H. Yang, Z. Yang,and J. Tang]{
    Xu Zou$^{12}$, Da Yin$^{12}$, Qingyang Zhong$^{12}$,  Ming Ding$^{12}$, Hongxia Yang$^{4}$, Zhilin Yang$^{*123}$,Jie Tang$^{*12}$
}
\affiliation{
    $^1$ Department of Computer Science and Technology, Tsinghua University
    \country{}
}
\affiliation{
    $^2$ Beijing Academy of Artificial Intelligence
    \country{}
}
\affiliation{
    $^3$ Recurrent AI, Ltd.
    \country{}
}
\affiliation{
    $^4$ DAMO Academy, Alibaba Inc.
    \country{}
}

\email{
  {zoux18,yd18,zqy20,dm18}@mails.tsinghua.edu.cn
}
\email{
yang.yhx@alibaba-inc.com.cn
}
\email{
   {zhiliny,jietang}@tsinghua.edu.cn
}

\renewcommand{\shortauthors}{Zou. et al}
\begin{abstract}
Large-scale pre-trained language models have demonstrated strong capabilities of generating realistic text. However, it remains challenging to control the generation results. Previous approaches such as prompting are far from sufficient, which limits the usage of language models.
To tackle this challenge, we propose an innovative method, inverse prompting, to better control text generation. The core idea of inverse prompting is to use generated text to inversely predict the prompt during beam search, which enhances the relevance between the prompt and the generated text and provides better controllability.
Empirically, we pre-train a large-scale Chinese language model to perform a systematic study using human evaluation on the tasks of open-domain poem generation and open-domain long-form question answering. Our results show that our proposed method substantially outperforms the baselines and that our generation quality is close to human performance on some of the tasks.\blfootnote{$^*$Corresponding authors: Zhilin Yang and Jie Tang.}

Narrators can try our poem generation demo at \url{https://pretrain.aminer.cn/apps/poetry.html}, while our QA demo can be found at \url{https://pretrain.aminer.cn/app/qa}. For researchers, the code is provided in \url{https://github.com/THUDM/InversePrompting}.




\end{abstract}

%
%


\maketitle

\section{Introduction}\label{sec:intro}


\begin{figure*}
    \centering
    \includegraphics[width=0.99\textwidth]{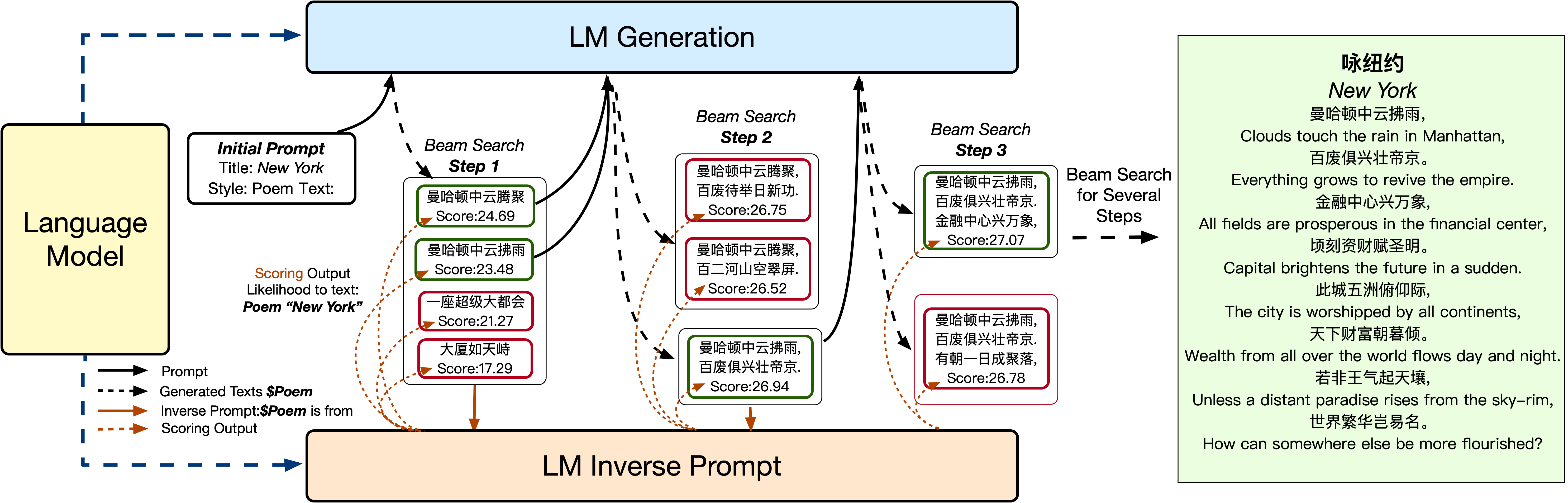}
    \caption{The generation process of open-domain traditional Chinese poems under \model. Using title \textit{New York} as an example. }
    \label{fig:newyork}
\end{figure*}

The field of text generation has made tremendous progress recently. Large-scale autoregressive Transformer models \cite{vaswani2017attention} optimized with maximum likelihood estimation have shown the ability of generating realistic text \cite{dai2018transformer,radford2018improving,brown2020language}. For real-world applications of text generation such as essay writing and story generation, it is essential for the users to be able to control the generation results. One of the most common approaches is to use prompting; i.e., a user shall manually write a few sentences to serve as the prompt and the language model generates the subsequent tokens given the prompt. For example, a user might input ``this is a sad story about a disease named COVID-19'' as a prompt to expect the generation of a COVID-19 story.

\begin{figure}
    \centering
    \includegraphics[width=0.5\textwidth]{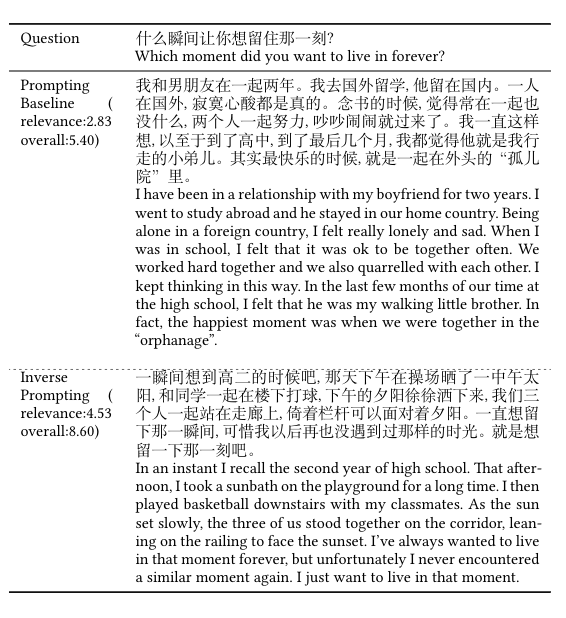}
    \caption{An example showing how the prompting baseline model may fail to maintain relevance in generated text, and how \model alleiates this issue. The relevance and overall scores were obtained from human evaluation.}
    \label{tab:example}
   \end{figure}

However, prompting is far from sufficient for controllable text generation. It is not uncommon for a language model to deviate the generation process from the original prompt and start generating text of unrelated topics. Figure \ref{tab:example} shows an example of how language models fail to maintain the coherence between the prompt and the generated text. In the example, the language model is asked to answer the question ``which moment did you want to live in forever''. The baseline using conventional prompting generates a story that deviates a lot from the prompt; i.e., most of the generated content is irrelevant to the question. There were also unnatural expressions that do not make much sense in the context.

To tackle this challenge, we propose a novel method, inverse prompting, to refine the process of text generation from pre-trained language models. Inverse prompting can be decoupled into three steps. First, given a piece of generated text, an inverse prompt is constructed using the generated text. Second, the conditional likelihood of the original prompt given the inverse prompt is computed based on the pre-trained language model. Third, the conditional likelihood is used as a score in beam search for selecting the best generation candidates. As a result, inverse prompting ensures that the pre-trained language model predicts the prompt given the generated text with high likelihood, which encourages the relevance of the generated text to the prompt. For example, in the case of question answering, a language model generates an answer given the question; during beam search, we use the same language model to compute the likelihood of the question given the generated answer for candidate selection. In the example of Figure \ref{tab:example}, text generated by inverse prompting describes a beautiful moment of high school times which is closely related to the question. As a result, inverse prompting achieves an average score of 8.60 out of 10 under human evaluation on this example, compared to the baseline's 5.40.


To systematically evaluate the performance of inverse prompting, we conduct an experiment using human evaluation on the tasks of open-domain long-form question answering and open-domain traditional Chinese poem generation. We pre-train a Chinese language model to serve as the base model in our experiments.
The task of long-form question answering is similar to answering questions on Quora or Zhihu. On this task, we show that inverse prompting achieves much higher scores in all aspects than the prompting baseline and the previous state-of-the-art Chinese language model CPM~\cite{zhang2020cpm}.
The task of traditional Chinese poem generation targets generating poems of an ancient form but with contemporary topics including rocket science, relativity, or artificial intelligence, which tests the generalization ability of different approaches. Figure \ref{fig:newyork} illsutrates an example of traditional Chinese poem generation under the title \textit{New York}. It combines contemporary notions of New York like Manhattan and the financial center with a traditional form and traditional poetic imagery of cloud and rain. On this task, human expert evaluation demonstrates that inverse prompting performs significantly better than the prompting baseline and is comparable to Jiuge~\cite{zhipeng2019jiuge}, a well-known state-of-the-art system for traditional Chinese poem generation. When we combine inverse prompting with self training, i.e., finetuning the model with self-generated poems, our system  outperforms Jiuge under human evaluation by a large margin.
Our results of human evaluation demonstrate that \model improves the controllability and quality of text generation significantly and achieves close-to-human results.

\section{Related Work} \label{sec:related}
\subsection{Pre-training and Language Models}

Language modeling has been widely used as an objective for pretraining and demonstrates strong generalization abilities. Originating from word embedding methods such as word2vec \cite{mikolov2013distributed} and GloVe \cite{pennington2014glove}, pretraining methods have displayed an increased level of importance in the field of natural language processing \cite{devlin2018bert,liu2019roberta,dai2018transformer}. These models are more general and require less domain-specific data to achieve strong performance.


Specifically, a main type of pretrained models are autoregressive language models. Generative pretraining (GPT) \cite{radford2018improving,radford2019language,brown2020language} and Transformer-XL \cite{dai2018transformer} achieve substantial improvement in terms of perplexity and also improves generation quality. The approach has also been adapted to different languages \cite{zhang2020cpm,de2020good}.



Although realistic text can now be generated automatically by large-scale pretrained language models, it is challenging but essential for users to be able to control the generation results. Prompting \cite{dai2018transformer,radford2018improving} has been widely used but is rather limited in controlling the generation results. CTRL \cite{keskar2019ctrl} proposes to use control codes to provide conditions for a language model. Different from their method, our method does not rely on modification of pretraining paradigms or human-designed attributes. PPLM \cite{dathathri2019plug} performs backpropagation during test time to adjust generation to maximize the scores given by attribute models. Compared to PPLM, inverse prompting does not require any gradient update to the original model and is free of any additional attribute models.

The idea of using dual process to strengthen the quality of AI generation by the dual property that the outputs and inputs are inputs and outputs under an inverse prespective has long been researched. ~\cite{xia2016dual} introduces dual learning for the task of machine translation. The method uses multiple different models to form a translation loop and hopes the contexts will remain unchanged after passing through the loop. CycleGAN~\cite{chu2017cyclegan} and VAE~\cite{an2015variational} also shares the similar idea of reconstruction in their applications.
Different from these works that uses different forward and inverse models, in this paper, we exploit the existence of inverse format in natural languages and use the same language model for prompting and inverse prompting. 


\subsection{Open-Domain Long-Form Question-Answering}

Question answering is a well-studied problem in artificial intelligence \cite{simmons1970natural}. There are various paradigms of question answering. Short-form question answering focuses on using a short phrase or sentence to answer the question \cite{rajpurkar2016squad,yang2018hotpotqa}. On the other hand, long-form question answering targets generating multiple sentences or paragraphs to answer a question in a more comprehensive way. Online question answering platforms such as Quora and Zhihu can be viewed as good examples of long-form question answering. While short-form question answering is easier to evaluate and more more widely studied, we are interested in investigate the ability of open-domain long-form question answering using text generation models in this work. Because it is challenging to evaluate the qualities of long-form question answering, we employ human evaluation in our experiments.



\subsection{Traditional Chinese Poem Generation}

Traditional Chinese poetry is an important genre of Chinese literature with a history of tens of centuries~\cite{liu1966art}. A few years ago, researchers experimented with generating traditional Chinese poems using statistical machine learning methods ~\cite{jiang2008generating}. Later, Jiuge~\cite{zhipeng2019jiuge,yi2020mixpoet} advanced traditional Chinese poem generation to a new level. As the well-recognized state of the art for open-domain Chinese poem generation, Jiuge is able to generate multiple forms of poems under any given titles, keywords or even images. 
Despite its ability to handle arbitrary open-domain inputs, Jiuge performs well on domain-specific contexts such as giant deserts or iron horses but does not generalize well to contemporary notions such as Donald Trump, quantum computation, and Europe. Different from Jiuge, we employ a large-scale language model pretrained on a general-purpose corpus and leverage inverse prompting to enhance its generation qualities.




\section{Methodology}\label{sec:problem}
In this section, we discuss the proposed \model method. The problem of text generation is modeled as generating $c_g$ given the prompt $c_p$, where both $c_p$ and $c_g$ are sequences of tokens.


\subsection{Baseline: Prompting and Beam Search}

Given a language model with probability distribution $p$, a simple and widely-used approach is to generate text by maximizing the conditional probability $p(c_g | c_p)$. This is usually achieved with beam search~\cite{medress1977speech}. With a beam size of $n$, beam search keeps the top-$n$ sequences during decoding time according to a beam scoring function $f(\cdot)$. An illustration is shown in Algorithm \ref{algo:beam}. The baseline method uses the log likelihood to define the scoring function, i.e., 
$f(c_g | c_p) = \log p(c_g | c_p)$


\begin{figure}
    \centering
    \includegraphics[width=0.46\textwidth]{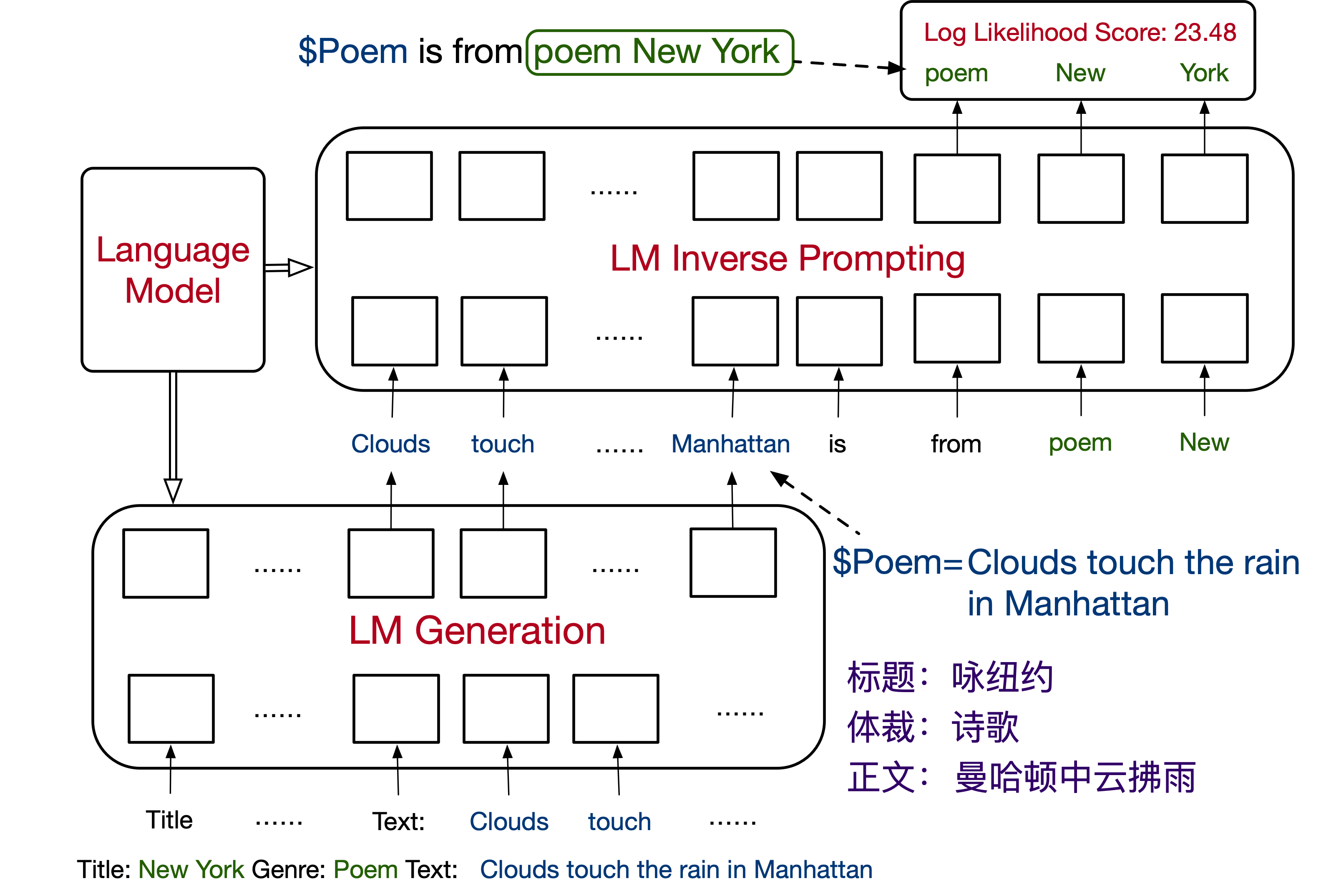}
    \caption{Language model generation and language model inverse prompting scoring for generating a poem sentence.}
    \label{fig:gip}
\end{figure}



\begin{algorithm}
\SetAlgoLined
\KwResult{Generated Context $c_{g}$}
 Given a language model $p$, a prompt $c_{p}$, the number of beams $n$, the number of steps $s$, exploration steps for each beam $m$. Initialize current step $k=0$. For each beam $j$, initialize the generated context for this beam $c_{j}=''$.
 
 \While{k<s}{
    For each $c_{j}$, generate $m$ next token $t_{j1}...t_{jm}$ sampled from $p(\cdot | c_p+c_{j})$, update $c_{jl}=c_{j}+t_{jl}$
    
    Choose the best $n$ texts $c_{b1},c_{b2}...c_{bn}$ with highest $f(c_{jl} | c_p)$.
    
    For all $j$, update $c_{j}=c_{bj}$.
    
    Update k=k+1.
    
 }
 Output the best beam $c_{g}=c_{1}$.
 \caption{Beam search. Inverse prompting follows the beam search framework with a novel scoring function $f$ being the inverse log likelihood of the prompt given the generated text.}
 \label{algo:beam}
\end{algorithm}







\subsection{Inverse Prompting}

In this paper, we introduce a new method based on a new scorer $f$ in beam search, called \textit{\model}.  Unlike previous controllable methods such as CTRL or PPLM which needs additional attribute model training or manually-defined codes, \model directly uses the original language model itself to improve its generation.

One main issue that reduces the quality of the generated text is the run-away phenomena shown in Table \ref{tab:example}. The text gradually becomes irrelevant to the given prompt as the sentences being generated. As the distance between the given prompt and the generated sentence becomes larger, it hinders the generator to keep a close connection with the prompt.

To alleviate this issue, our main idea is to design a novel beam search scoring function that evaluates the log likelihood in an inverse direction; e.g., if the prompt can be generated back from the text, they ought to be very related with each other:
\beq{
\label{eqn:inv-1}
f(c_{g} | c_p)=\log p(c_p|c_{g}).
}

Texts are not always fluent if we read them from an inverse way. In question-answering, the prompt may be "Question:\textbf{\$\{Question\}} Answer:". It is natural to follow the answer after that, yielding "Question:\textbf{\$\{Question\}} Answer:\textbf{\$\{Answer\}}". However, it is very unlikely that in natural language the order will present in the inverse way  "\textbf{\$\{Answer\}} Question:\textbf{\$\{Question\}} Answer:". Simply using equation \ref{eqn:inv-1} only results in failure.

However, thanks to the nature of natural language, there do exist ways to rearrange contexts to make them appear in an inverse order properly. Let's continue with the above instance: For "Question:\textbf{\$\{Question\}} Answer:\textbf{\$\{Answer\}}", there do exist a way in natural language to inverse it: "\textbf{\$\{Answer\}} answers the question:\textbf{\$\{Question\}}". 

\begin{figure*}[t]
   \centering
   \includegraphics[width=\textwidth]{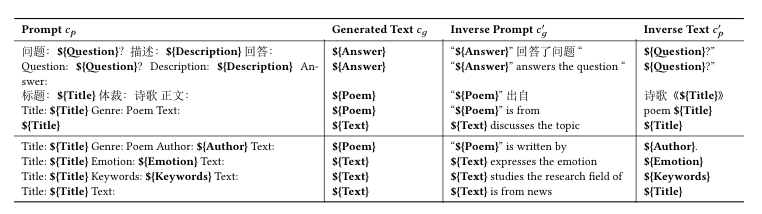}
   \caption{\label{tab:inv} Inverse prompting transformation Table. The first rows represents the inverse prompts used in experiments.(in Chinese and English) Some additional examples of \model format are also displayed. }
  
\end{figure*}

To achieve the core idea of Eqn. ~\ref{eqn:inv-1}, we simply need to alter the format of the prompts and texts:
\beq{
\label{eqn:inv-2}
f(c_{g} | c_p)=\log p(c'_p|c'_{g}),
}
where $c'_{g}$ is inverse prompt under a new format, and $c'_p$ being the inverse text. Figure \ref{tab:inv} displays some examples of this transformation format.  For $c_p=$"Question:\textbf{\$\{Question\}} Answer:" and $c_{g}=$\textbf{\$\{Answer\}}, we only need to set $c'_p=$"\textbf{\$\{Question\}}" and $c'_{g}=$"\textbf{\$\{Answer\}} answers the question:",  equation ~\ref{eqn:inv-2} shall work. \modelst ranks different beams by their likelihood to generate back the original prompt in an inverse way, promoting the most relevant generated texts. 
\modelst can be used as long as the language supports an inverse structure to rearrange the prompt and context in a proper way. Detailed illustration for language model generation and language model \model is presented in Figure \ref{fig:gip}. 

The top rows in figure \ref{tab:inv} are the formats of \model in the two experiments of our paper---long-form question answering and poem generation, while the bottom rows are additional examples of how \model can be used in other text generation tasks.

\modelst is a simple method and easy to implement. The method requires no additional models or data processing, as the inverse prompting score can be simply computed by the same language model used for generation. However, \model offers large improvements to the qualities of the generated texts, which we will show in Sections \ref{sec:model} and \ref{sec:exp}. 

\section{Implementation}\label{sec:model}

We mainly use two long-term text generation tasks, Open-Domain Long-Term Chinese Question-Answering, and Open-Domain Chinese Poem Generation, which require the AI to generate long, in-depth contexts according to relatively short prompts, to demonstrate the fantastic performance of \model. 

We believe that as the relevance between generated texts and the given prompt (questions/titles) improves, the generation quality will increase too. So we conduct \model on questions/titles in our experiments, as shown in the first four rows in Figure \ref{tab:inv}.

\subsection{Base Language Model}

We train our base Chinese language model using Megatron-LM\cite{shoeybi2019megatron} with Transformer-XL\cite{dai2018transformer}. The model has 2.86 billion parameters. The training set contains 302GB of raw Chinese data abstracted from multiple Chinese websites including Baidu, Zhihu and Sougou. We train the base model using 64 GPUs for 160,000 steps. Details of training settings and datasets are displayed in Appendix~\ref{app:detail}. 

\subsection{Open-Domain Long-Form Question-Answering}

Long-Form Question-Answering, like the QAs on Quora, Zhihu or Sougou, is a form of question-answering that questions and descriptions are short and described in one or two sentences, while the answers have to be long, informative and in-depth. The questioner expects a thorough answer answering the question he asks in detail. 

We apply \model in this way to generate Long-Form Answers given Question prompts. We generate sub-sentences randomly according to language model $\mathcal{LM}$, and do beam-search with \model in sub-sentence level. To ensure the answer follows the question, we apply \model (Equation \ref{eqn:inv-2} for each sub-sentence and sum up their scores. To keep the generated context fluent, we combine the scores with normalized forward perplexity,
 
\beq{
\label{eq:inv}
f(c_{g}|c_{p})=\frac{1}{n}\sum_{s\in c_{g}} \lambda_1 \log p(c'_{p}|s')+\lambda_2\frac{\log p(c_g|c_p)}{n(c_g)^\lambda}.
}

\subsection{Open-Domain Poem Generation}

Traditional Chinese Poem generation is the pearl of domain-specific long-form Chinese text generation tasks. Traditional Chinese poems have their specific complex format and word usages different from modern Chinese language. Most of the poems are written by ancient poets to express their feelings, describe things they are doing, or ideas on different items. Generation of meaningful text under the poem format given open-domain information is very hard for both state-of-the-art AI models and humans.

In this paper, besides Open-Domain Long-Form QA, we challenge our \model for a seemingly impossible task-- To use the language model trained on modern texts to generate Open-Domain Traditional Chinese Poems. 

We basically keep the \model format of equation \ref{eq:inv} while adding a poem-format term to the beam-search (Equation \ref{eq:poemgen}), which penalizes contexts by the degree they disobey with the poem format in rhythms or tones. 
\beq{
\label{eq:poemgen}
f(c_{g}|c_{p})=\frac{1}{n}\sum_{s\in c_{g}} \lambda_1 \log p(c'_{p}|s')+\lambda_2\frac{\log p(c_g|c_p)}{n(c_g)^\lambda}-\lambda_3 l_{format}(c_g)
}
\subsection{Self Training for Poem Generation}
\label{sec:reinforce}

Given that the model is trained on modern Chinese texts including very few poem-format texts, it can hardly generate texts fully obeying the poem format while maintaining a strong relevance to the given title. 

Therefore, to improve its performance, we try the generate-and-fine-tune self training protocol in AlphaGo-Zero\cite{silver2017mastering} for this task. 

We randomly select 1500 titles and let the model to produce poems based on them. Then we fine-tune the model on these generated poems for 2000 steps. This cycle can be repeated multiple times and in our experiments we repeat this cycle for 2 times. We expect the fine-tuned model to be more likely to generate sentences with better poem formats and other poem-specific properties like aesthetics without losing their relevance to the given title.

\section{Experiments}\label{sec:exp}

In this section, we display the human-evaluation results of \model on two long-form text generation tasks, open-domain long-form QA and open-domain poem generation.

\subsection{Human Evaluation Protocol}
We first introduce how our human evaluation on the two tasks is conducted. 
Table \ref{tab:humaneval} illustrates the statistics for human evaluation experiments. For open-domain long-form QA, we recruit 45 people, mostly university students, to evaluate the quality of the generated answers.
As for the evaluation of poem generation, we invite 11 experts on traditional Chinese poems. Some of them previously participated in the evaluation of Jiuge, the previous state-of-the-art Chinese poem generator. The others are either similarly known to be knowledgeable on poems or guaranteed to be qualified for the evaluation by some poem contests.

\begin{table}[ht]
\small
    \centering
    \caption{Human Evaluation Statistics. We filter out the answers from participants who does not finish all questions and pass the consistency check.}
    \label{tab:humaneval}
    \begin{tabular}{p{0.17\textwidth}rrrr}
    \toprule
         \multirow{2}{*}{Task} & Participants & Prompts &Methods & Scores \\
         &   & &Compared&Collected \\
         \midrule
         long-form QA & 30   & 100&4&12,000 \\
         Poem generation & 10 & 100 &4&4,000\\
    \bottomrule
    \end{tabular}
\end{table}
Each task contains 100 prompts and for each prompt, we provide 4 different contexts for evaluators to evaluate. An evaluator needs to score each context from multiple aspects and give an overall rating for each context on our online evaluation platform within one week time. 

To ensure participants making evaluations seriously, for each prompt we ask the participants to select the best context. Then we will check if this answer is consistent with the overall ratings for those 4 contexts additionally. If the proportion of inconsistent answers reaches over 20\%, we will treat this participant as invalid. Finally, we only collect the answers submitted by valid participants.

As listed in Table \ref{tab:humaneval}, 32 evaluators in long-form QA evaluation and 10 experts for traditional Chinese poems finished his/her evaluation. 30 of the finished evaluators in long-form QA experiment are valid, while all the 10 finished experts in our traditional Chinese poem experiment are valid.

\subsection{Open-domain long-form Chinese QA}

For open-domain long-form Chinese QA evaluation, we randomly select 100 questions from various domains in Zhihu, a Quora-like Chinese open-domain long-form QA platform. In Zhihu, users can ask questions with some background descriptions, or provide informative long-form answers to the raised questions. Besides, users can "Upvote" or "Downvote" answers based on their opinions.

In this experiment, we only select questions that are excluded in the training set of our base model. For each question, we display one highly-upvoted human answer and three AI-generated answers produced by CPM~\cite{zhang2020cpm}, prompting baseline, and \model respectively. 

We shuffle the order of all answers and ask human evaluators to score the answers through four aspects including:
\begin{enumerate}
    \item \textbf{Fluency} Whether the answer is well-formed and logical to read. Rated from 1 to 5.
    \item \textbf{Informativeness} Whether the answer contains useful information for the given question. Rated from 1 to 5.
    \item \textbf{Relevance} Whether the answer is relevant to the given question. Rated from 1 to 5.
    \item \textbf{Overall} The overall quality of the answer. Rated from 1 to 10.
\end{enumerate}


\begin{table}[ht]
\small
    \centering
    \caption{Performance for open-domain long-form Chinese QA under Human Evaluation. }
    \label{tab:open-domain-qa-human-eval}
    \begin{threeparttable}
    \begin{tabular}{lccccc}
    \toprule
         \multirow{2}{*}{Method} & Fluency & Inform.\footnotemark[1] & Relevance & Overall \\
         & (1-5) & (1-5) & (1-5) & (1-10) \\
    \midrule
       CPM~\cite{zhang2020cpm} &2.66&2.47&2.36&4.32\\
        Prompting Baseline &3.44&3.25&3.21&5.97 \\
        Inverse Prompting &\textbf{3.61}&\textbf{3.43}&\textbf{3.59}&\textbf{6.51} \\
    \cmidrule{1-5}
        Human Answers &3.80&3.61&3.67&6.85\\
    \bottomrule
    \end{tabular}
    \begin{tablenotes}
        \item[1] Informativeness
    \end{tablenotes}
    \end{threeparttable}
\end{table}

Table~\ref{tab:open-domain-qa-human-eval} shows that \model outperforms both the prompting baseline and the previous SOTA Chinese language model CPM by a large margin in all individual aspects, as well as the overall quality of the generated answers.

Despite \model only forces the answer to be more related to the question in this experiment, an interesting finding is that by producing more relevant answers, \model also makes improvements on the fluency and informativeness of the generated answers, raising the overall quality as well. This supports our hypothesis in section \ref{sec:model}.
\subsection{Open-domain Poem Generation}

The second experiment is to evaluate the task of open-domain poem generation. This task is similar to the long-form QA experiment described above. We randomly design 100 poem titles including various open domains for evaluation. These poem titles never appear in any real poems in the training set or being used as randomized titles in our reinforcement learning process.


For each title, we apply four different methods to generate pomes, including Jiuge (the SOTA model for open-domain Chinese poem generation), the beam search baseline with poem format loss $l_{format}$, \model with poem format loss (Equation \ref{eq:poemgen}) and \model with the self-training mentioned in section \ref{sec:reinforce}.
These four poems are shuffled for evaluation. For each generated poem, we request evaluators for 5 ratings:
\begin{enumerate}
    \item \textbf{Format} Whether the generated poem follows the rule of rhythm in traditional Chinese poetry. Rated from 1 to 5.
    \item \textbf{Innovation} Whether the sentences are copied from existing poems or created with innovative expressions. Rated from 1 to 5.
    \item \textbf{Relevance} Whether the content of the poem is related to the given title. Rated from 1 to 5.
    \item \textbf{Aesthetics} Whether the poem has obscure meanings apart from its obvious meanings, making it aesthetically better? Rated from 1 to 5.
    \item \textbf{Overall} The overall quality of the poem. Rated from 1 to 10.
\end{enumerate}
\begin{table}[ht]
\footnotesize
    \centering
    \caption{Performance for open-domain Traditional Chinese Poem Generation under human expert evaluation.  }
    \label{tab:poem_eval}
    \begin{threeparttable}
    \begin{tabular}{lccccc}
    \toprule
         \multirow{2}{*}{Method} & Format & Innov.\footnotemark[1] & Relevance & Aes.\footnotemark[2] & Overall \\
         & (1-5) & (1-5) & (1-5) & (1-5) & (1-10) \\
    \midrule
      Jiuge~\cite{zhipeng2019jiuge} &\textbf{3.60}&2.47&1.99&\textbf{3.12}&3.57\\
        Search Baseline &2.79&1.10&1.16&2.44 &1.35  \\
    \cmidrule{1-6}
        Inverse Prompting & 2.56&2.71&2.92&2.33&4.00  \\
        Inverse Prompting +ST & 2.42&\textbf{2.92}&\textbf{3.65}&2.18& \textbf{4.40} \\
    \bottomrule
    \end{tabular}
    \begin{tablenotes}
        \item[1] Innovation
        \item[2] Aesthetics
    \end{tablenotes}
    \end{threeparttable}
\end{table}

 Table~\ref{tab:poem_eval} illustrates the experimental results. The average scores for all methods are low as all of the experts are extremely critical. They only give high scores to very few perfect poems. One of the experts says she'll give less than 5 "overall" score to an average TC-Poem written by ancient celebrities, while scoring results indicate that other experts are even more critical than her.  
 
 The prompting baseline can hardly generate appropriate poems. Even with the poem format loss, it only outputs unrelated sentences copied from existing famous poems that appear in modern Chinese languages. 
 
 However, with the help of \model, the overall quality of generated poems surpasses \textit{Jiuge}. Moreover, the self-training can further improve the performance on top of \model.

Generally, \textit{Jiuge} is good at generating sentences with beautiful words and gorgeous rhythm, since it is designed to focus strictly on poem formats. Nevertheless, according to human evaluation results, despite it sometimes does generate relevant poems, most of its generation results are not quite related to the given title and comparably weak at innovation.  

 \model offers innovative and relevant expressions in the form of traditional Chinese poems. However, as the model is trained on modern Chinese texts, it is inevitably not so good in following traditional poem formats like rhythms or tones. It also doesn't handle aesthetics well, as this is common in ancient Chinese poems, but rarely appears in modern Chinese.

Despite these disadvantages, the experts still agree to give poems generated by \model a much higher average overall score than Jiuge due to their high relevance to titles and innovative expressions. 

In section \ref{sec:reinforce}, we expect the self-training can bring better format and aesthetics to the generated poems. However, to our surprise, the self-training further enhances the innovation and relevance by a large margin at the cost of a minor decrease in format and aesthetics, suggesting that what the model really learns under our reinforcement learning scheme is to be more relevant. By generating more relevant poems to the title with more innovative expressions, its average overall score becomes much higher additionally. Eventually, \model with self-training gets 4.40 average overall score, compared with Jiuge's 3.57.

One possible explanation for this phenomenon is that in order to be more relevant to open-domain titles which may never appear in the domain of Traditional Chinese Poems, the model has to be more innovative in language arrangement and less focused on formats or aesthetics. 

In Appendix \ref{app:p-value}, we discuss our deviation analysis and calculate p-values for different methods on the above two tasks. 


\subsection{Poem Turing Test}

Apart from human evaluation for open-domain titles, we also test the performance of it on domain-specific titles. 

Back to the result of long-form QA in Table~\ref{tab:open-domain-qa-human-eval}, answers generated by \model are only slightly inferior to human answers. Their average score is 6.51 compared with human answers' 6.85.  This enlightens our mind that the generated poems may be comparable in quality to human-made poems. 

Inspired by turing test~\cite{turing2009computing}, we similarly designed a traditional Chinese poem turing test to further evaluate the generated poems quality of \model.

In the turing test, also known as the imitation game, a human interrogator is requested to distinguish between generated poems and human poems. We implement an online game platform
where any player can participate without limitation. In the game, each player is given several pairs of poems with each pair contains one poem written by a human poet and the other one generated by AI under the same title. The human poems are randomly selected from \textit{Quan Tang Shi}, the most famous collection of traditional Chinese poems. The collection was published in 1705 and consists of high-quality traditional Chinese poems mostly written in Tang Dynasty (618-907).
In our designed game, the player needs to figure out which poem is written by the human poet. We generate 1,500 pairs of poems and randomly displays 5 pairs for each game. 

As displayed in Table~\ref{tab:turing}, 4,592 game records are gathered from 755 different users. Each game record involves a binary selection between a human poem and an AI poem. 46.4\% of the user records choose AI poems while the rest 53.6\% chooses human poems, suggesting that the quality of poems generated by \model+self-training on domain-specific titles may be close to human level for average online users.

\begin{table}[ht]
    \centering
    \caption{Poem turing test on 755 identical online users. Human poems are randomly selected from \textit{Quan Tang Shi}. AI Poems are generated given the same title.}
    \label{tab:turing}
    \begin{tabular}{lrrr}
    \toprule
         Method & Total & Selected& Selection Rate \\
    \midrule
      Inverse Prompting +ST& 4,592 & 2,132 & 46.4\%\\
     Ancient Human Poems &4,592 & 2,460 & 53.6\%  \\
    \bottomrule
    \end{tabular}
\end{table}
\subsection{Case Study}
In this section, we discuss explicitly the pros and cons of \model by performing series of case studies on the two tasks -- Open-domain long-form QA, and Open-domain Poem Generation. 
\begin{figure}[ht]
    \footnotesize
    \centering
    \includegraphics[width=0.5\textwidth]{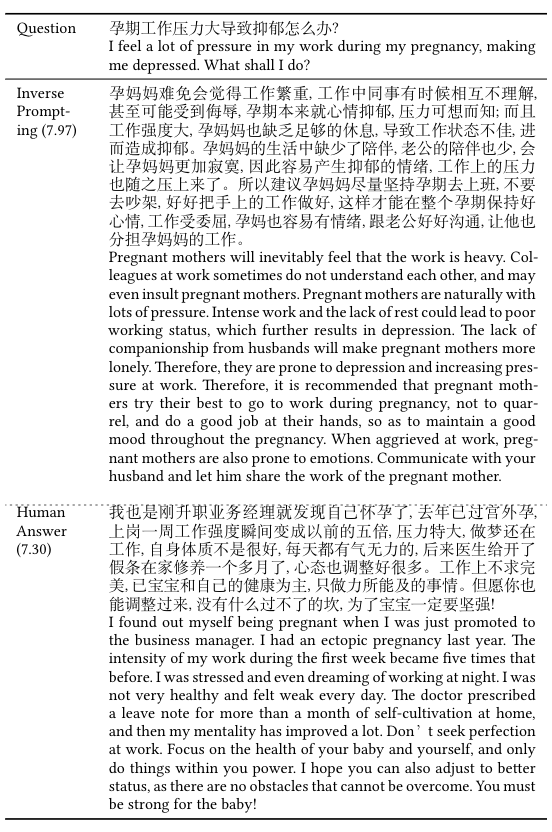}
    \caption{A Perfect Example of \model generating better answer than human in open-domain long-form QA.}
    \label{tab:qa-answer-pregnant}
   
\end{figure}

Figure~\ref{tab:qa-answer-pregnant} exhibits a comparison between two answers for a question on how to deal with stress at work during pregnancy. We list the answer generated by \model and the human answer. In this case, the evaluators even give higher scores to the \model generated answer than the human answer. Both answers provide comprehensive and practical aids related to the question. The human answer is more emotional and gives advice based on the answerer's own experience. The generated answer, on the other hand, raises the point that pregnant mothers should insist on working and overcome the difficulties straightforwardly. Then it uses several sentences to enhance this point, which turns out to be more informative, reasonable and persuasive.

\begin{figure}[ht]
    \footnotesize
    \centering
    \includegraphics[width=0.5\textwidth]{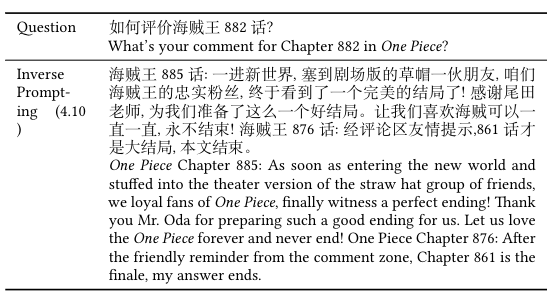}
    \caption{A bad case for \model generated texts. It can't overcome the barrier of maths.}
    \label{tab:qa-answer-onepiece}
  
\end{figure}

While the proposed method seems to be able to understand questions, integrate learned knowledge and generate logical answers, we found that numbers in the task are comparatively difficult, which often lead to some chaotic outputs. In Figure~\ref{tab:qa-answer-onepiece} we show a bad case generated by \model that only receives a 4.10 score in overall quality. While the question is asking about Chapter 882 in the \textit{One Piece} manga, the model is clearly unable to understand the differences between Chapter 885 and the asked 882. Besides, the answer itself is self-contradictory. It is worth noticing that such a chaotic problem in maths universally exists for language models. Previous study~\cite{Saxton2019AnalysingMR} also shows that it is extremely hard for neural models to do mathematical reasoning.

\begin{figure}[ht]
    \footnotesize
    \centering
    \includegraphics[width=0.5\textwidth]{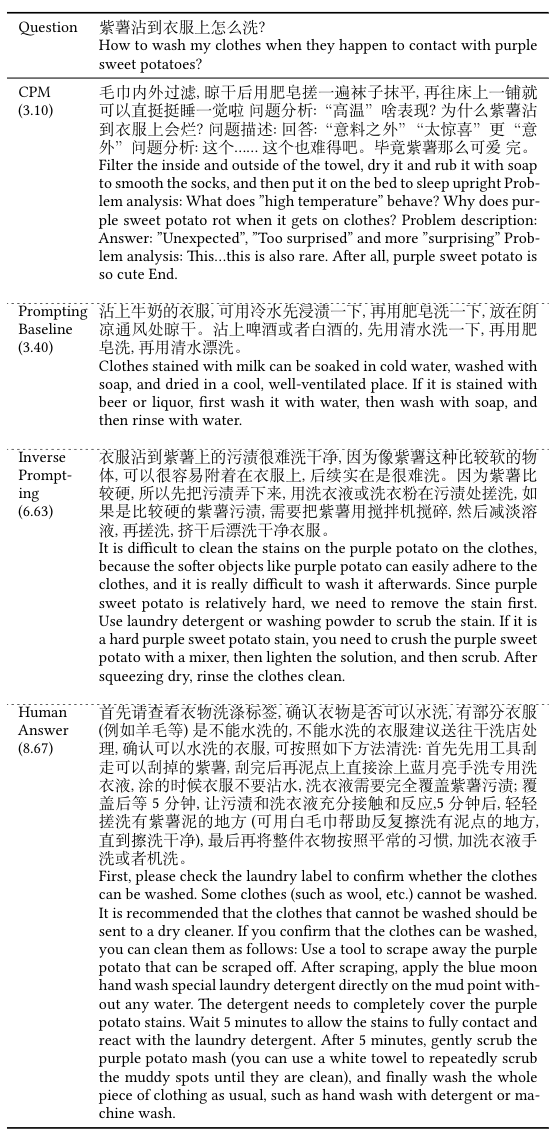}
    \caption{A representative case in open-domain long-form QA that the quality of answers in this problem reflects the overall performance of different methods. }
    \label{tab:qa-answer-wash-cloth}
    
\end{figure}

In Figure~\ref{tab:qa-answer-wash-cloth}, we display all 4 answers for the question ``How to wash purple sweet potato on clothes'' with the average overall scores. The best answer is written by a human, which comprehensively introduces the solution to the problem step by step. The answer generated by \model offers a similar solution but with fewer details. The prompting baseline does not give a precise answer to the original question. Instead, it tries to answer another question ``How to wash out the liquids on clothes such as milk or beer?''. This tells us why we need to use \model to force the generated answer to be closely related to the original question. Finally, CPM can neither produce fluent answers nor provide useful information. This example illustrates how the difference in overall ratings for different methods in Table~\ref{tab:open-domain-qa-human-eval} come from in a representative way.

\begin{figure}[ht]
    
    \centering
    \includegraphics[width=0.5\textwidth]{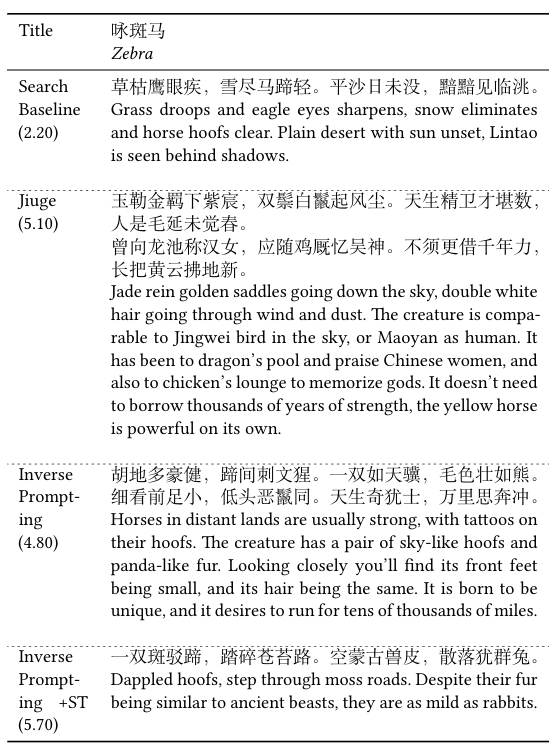}
    \caption{A representative case in poem generation that reflects the overall qualities of different methods. }
    \label{tab:tcp-zebra}
    
\end{figure}

\begin{figure}[ht]
    \centering
    \includegraphics[width=0.5\textwidth]{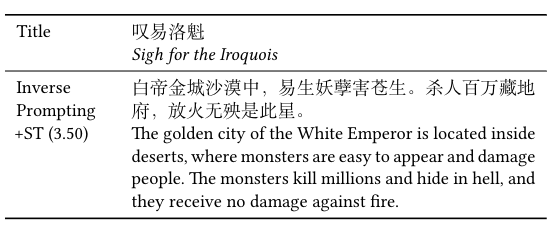}
    \caption{A bad case that \model is not successful in understanding the given title Iroquois even under after self training. }
    \label{tab:tcp-bad}
  
\end{figure}

\begin{table}[ht]
    \small
    \centering
    \caption{A GPT-2 345M paragraph generation case. }
    \label{tab:tcp-eng1}
    \begin{tabular}{p{0.05\textwidth}p{0.38\textwidth}}
    
    \toprule
        Title& Arctic Cat  
        \\
        \midrule
        GPT-2 &VIRGO (Aug. 23-Sept. 22): A business partner invites a group of friends to a casino but no money is exchanged, causing a misunderstanding. A lawyer accidentally chases away a fish when someone else asks for the name of a housemate.
        
LIBRA (Sept. 23-Oct. 22): A business partner mentions missing money on a Web site, triggering a conversation about receiving the money in time. Expect to be caught by someone who knows the truth.

SCORPIO (Oct. 23-Nov. 21): When others make big bets, you have to say no. If you don't like it, they can make it worse. But don't do it.

SAGITTARIUS (Nov. 22-Dec. 21): Take a stand and speak your mind. Don't let others tell you you can't do it. You can.

CAPRICORN (Dec. 22-Jan. 19): Speaking too much may seem harsh, but it's important. For example, if someone tries to take over your career, you must not cave in.

AQUARIUS (Jan. 20-Feb. 18): You are correct to think that everyone agrees with you when you talk about this topic. However, this person may not be a professional so don't break bread with them.

PISCES (Feb. 19-March 20): When someone tells you they're really sorry for throwing away something, it is OK to say yes. Keep it short and simple.

IF APRIL GOES AWAY

Aries (April 20-May 20): You are able to talk easily with someone who is far away from you because of a move of public events.

Taurus (May 21-June 20): If you trust someone, you might get involved with them by going out with them.

GEMINI (June 21-July 22): Don't act your age when you are with someone and they get angry. You are mature enough for such a thing.

CANCER (July 23-Aug. 22): You can take on anything in life that you like, but others don't like it. That is OK.

Leo (Aug. 23-Sept. 22): They see you as someone they can manipulate and can be loud about. Don't like it.

\\

\midrule
Inverse Prompting & 
Around the globe, it's still hard to find a cold-weather car. However's will now serve as the closest thing to snow on the road yet. The new Arctic Cat Aruba features two levels of navigation that allow drivers to get a glimpse of location and what's happening on the road. Additional intercom-compatible pieces will be available to start the journey when it makes its return to the United States. Further out Arctic Cat Aruba For the US.
\\
    \bottomrule
    \end{tabular}
\end{table}

\begin{table}[ht]
    \small
    \centering
    \caption{Another GPT-2 345M paragraph generation case. }
    \label{tab:tcp-eng2}
    \begin{tabular}{p{0.05\textwidth}p{0.38\textwidth}}
    
    \toprule
        Title& Boris Johnson in Delhi 
        \\
        \midrule
       GPT-2 & And far from showing relief that the switch-back from the hard-nosed - and potentially racist - EU-UK relations to a softer, conciliatory approach, Tony Blair yesterday morning raised expectations of another barrage of retaliatory tariffs against the UK's soaring exports to America, and said he was in no mood to settle for the kind of bilateral exchange once favoured by George W Bush and Tony Blair.
        
Downing Street has been irritated that the free-trade deal with America was unexpectedly allowed to lapse just when the G7 summit was over - something it had tried hard to avoid.
But the prime minister, in a direct intervention before the G7 summit in Northern Ireland, suggested the long-awaited deal would take time to complete.

He told Sir Michael Dawe: "We want to get to the point where we are able to settle trade in terms of good terms with the United States on a relationship that is not just good for the United States, but is good for the UK as well.

"It is an old parochial argument that trading relations are too important to be put at the mercy of political relations. That's a fundamentally flawed argument."

Mr Blair has faced particular criticism from the Tory backbenchers over the continued access of the US and British airlines to market dominance through the "free movement" of goods, which has now been abandoned and which Mr Blair publicly admitted as a feature of the global economy.

That prompted the shadow trade secretary, Chris Grayling, to accuse the prime minister of trying to impose unilateral change on the economic process.

In a letter to Mr Grayling, he said: "It seems that for the moment, Mr Blair is pushing for an extra ten days in the EU deal for our exports to America without consulting us. The UK has been up against that earlier history in a number of ways - with the currency exchange rate as well as the fact that exports and imports have been flowing at cross purposes.

"But there are some substantive things that Mr Blair might be looking at that we need to have on our side, and that would be an effort to get these and other issues sorted out before the G7 summit. It would probably be a necessity before the G7 summit, before this financial rescue package."

The Labour leader, Harriet Harman, added: "The proposed U.S.-UK free trade agreement that the prime minister has opposed is going to fail if we don't deal properly with it.
\\

\midrule
Inverse Prompting & 
May 24, 2009 - Boris Johnson arrives at the National Portrait Gallery to accept a prize for his portrait of Italian Chancellor Romano Prodi

July 28, 2011 - Boris Johnson makes his first visit to south Asia, in the Indian capital, Delhi

June 4, 2012 - Boris Johnson attends a US delegation in New Delhi

June 20, 2013 - Boris Johnson visits the Taj Mahal monument in Agra, the Raj, Bihar holy city of Lalbagh

November 21, 2014 - Boris Johnson in his office, New Delhi

January 23, 2014 - The foreign with which Boris Johnson has most bonded during the ten-month tour of India and China was Prince Charles. He met the billionaire at Buckingham Palace before heading off to Delhi.
\\
    \bottomrule
    \end{tabular}
\end{table}

Figure \ref{tab:tcp-zebra} shows poems generated by different methods under title \textit{Zebra} . Zebra is an open-domain concept that never appears in any traditional Chinese poems (as there's no zebra in ancient China). However, there exist lots of traditional Chinese poems for different types of animals. We would like to see how different methods generalize the traditional Chinese poem format for zebras.

Note that the direct meaning for ``zebra'' in Chinese is "spotted horses", so models are likely to misuse the concept of "horse".
The prompting baseline copies sentences from famous existing poems for horses and gets only 2.20 for being an awkward copycat.  Jiuge treats zebras as horses and applies a lot of analogy to glorify the "horses", with good representation and perfect format it gets a 5.10 overall score.  \modelst offers a description between horses and zebras and gets a 4.80 overall score, while \model with self-training states the hoof, the fur and the behaviors of zebras in a subtle way, differing zebras from horses, this poem is scored the highest (5.70) by expert evaluators.

This is a representative instance for poem generation. The other 99 poems are also evaluated in such a way that the title is in a category that exists a lot in traditional poems. However, the precise concept of the title is new. For example, climbing different mountains or towers is popular in traditional Chinese poems, and we design open-domain titles like ``Climbing Eiffel Tower'' and ``Climbing Mt. Everest'' which never appear in them. The prompting baseline often copies from existing poems. Jiuge usually gives an poem with good format between related and unrelated. \model seems to understand the title while \model with self-training understands the title better. 

However, \model does not guarantee understanding of the given title. Figure~\ref{tab:tcp-bad} illustrates a bad case under title \textit{Sigh for the Iroquois} . \modelst+self-training fails to understand the meanings of the native American tribe and mistreat it as some forms of monsters living in deserts and killing people. This may due to the low occurrence of the Iroquois in modern Chinese texts that the base language model itself cannot fully understand its meanings, and the self-training process is unable to help the model understand more about such concepts as well.

\subsection{Inverse Prompting for English(GPT-2)}

We also practice inverse prompting under an open-sourced toy English language model: GPT-2 345M\footnote{\url{https://github.com/NVIDIA/Megatron-LM}}\cite{shoeybi2019megatron}. 

As can be concluded from Table \ref{tab:tcp-eng1},\ref{tab:tcp-eng2}, inverse prompting greatly improves the relativeness of GPT-2 345M for generated English contexts. Under title "Arctic Cat", inverse prompting refers to a cold-weather car brand, while direct generation generates totally unrelated contexts. Under title "Boris Johnson in Delhi",GPT-2 is successful in referring the title to UK, while completely ignoring "Delhi". However, using inverse prompting, the generated context suddenly becomes very relative. 

Although the quality of the base model limits the performance, in these cases inverse prompting still achieves obvious improvements. 

\section{Conclusion} \label{sec:conclusion}
In this paper, we present a new method, \model for text generation. \model offers a new option for controllable generation using language models by exploiting the inverse form of natural languages. 

We experiment the text generation using \model, which is to assign inverse prompts that forces the generated texts to be related to the prompt. Extensive experiments of human evaluations demonstrate that as the relevance increases, the overall quality of long-form Chinese texts also improves significantly. On long-form open-domain QA, \model improves AI-generated texts one step closer towards human level. More promising results occur in open-domain Traditional Chinese Poem Generation, as with \model, language models trained on modern Chinese context can generate poems that surpass the previous SOTA despite the disadvantages in format and aesthetics. Furthermore, with self-training, the \model poem generator can do even better in human evaluations.

Future works may include improving the poem generation format, applying \model to other languages like English, or enhancing the self training-process. 

\vpara{Acknowledgements}

We thank Zhengxiao Du and Jifan Yu for helpful discussions. 

Ming Ding contributed for the speed acceleration after submission to KDD. So we decide to list him in the author list of the arxiv version.

\section{Appendix}
\subsection{Implementation Details}
\label{app:detail}
\vpara{Training Details for Base Model.}
For training of base model, we use a training set of 302GB, the distribution of these data is shown in Table~\ref{tab:trainset}. The evaluation set contains 400MB Open-Domain QA contexts that is not used during training. We select the 100 questions in human evaluation from this evaluation set.
\begin{table}[ht]

    \centering
    \caption{Dataset Distribution.}
    \label{tab:trainset}
    \begin{tabular}{llr}
    \toprule
         Source & Format & Size \\
    \midrule
      Baidu \& Sougou Baike & Online Encyclopedia & 133GB\\
      Zhihu & Open-domain QA & 131GB \\
      Baidu QA & Open-domain QA & 33GB \\
        Generated TC-Poems & Traditional Chinese Poetry & 1.6MB\\
     \midrule   
       Evaluation & Open-domain QA & 440MB \\
    \bottomrule
    \end{tabular}
\end{table}

As mentioned in section~\ref{sec:model}, we use GPT framework with its transformer model substituted to Transformer-XL. For optimization, we use the AdamW optimizer with $\beta_1=0.9,\beta_2=0.95, \epsilon=1e-6$ and a $0.1$ L2-weight decay. The learning rate is warmed up linearly over the first 3,000 steps to a peak value of $1e-4$, then is tuned with cosine decay to $10\%$ of its peak value. The total training steps is 160,000. The training process uses 8 servers with 8 Nvidia V100 GPUs on each server. Each server has 96 Intel CPU cores and 376GB Memory. Serves are connected by 100G RoCEv2 network. 

For reinforcement learning, on each cycle we first generate a few poems for each of the 1500 prompts, resulting in around 800KB of generated poem data. Our fine-tuning inherits the previous conditions of the optimizer from the previous model and train on generated poem data for 2,000 steps. We repeat this process twice, so the final size of train poems generated is 1.6MB. The fine-tuning uses one server with 8 Nvidia V100 GPUs.  

\begin{table}[ht]
    \small
    \centering
    \caption{Training Dataset Distribution.}
    \label{tab:beamsearch}
    \begin{tabular}{p{0.18\textwidth}rrr}
    \toprule
         \multirow{2}{*}{Task} & beam & generations  & max short \\
         & size &  per beam &  sentences \\ 
    \midrule
      Long-form QA &  5  & 5   &  30 \\
      Poem (train/turing) & 10    & 7  & 8\\
        Poem (eval) & 10 & 12 & 8\\
        \bottomrule
    \end{tabular}
\end{table}

\vpara{Parameters for Beam Search.}
Table~\ref{tab:beamsearch} displays the beam search parameters we use. For long-form QA, we use a beam size of 5, and for each beam we generate 5 samples for the next short sentence, and we limit the length of the answer to 30 short sentences. For Poem Generation, we use a beam size of 10, for each beam we generate 7 samples for the next short sentence in reinforcement learning and the Turing Test, and 12 samples for open-domain title human evaluation. We limit the length of the generated poems to 8 short sentences. 

For $\lambda,\lambda_1,\lambda_2,\lambda_3,l_{format}$ mentioned in Section~\ref{sec:model}, we take $\lambda=\lambda_1=\lambda_3=1$, $\lambda_2=0.75$ for poem generation and use $\lambda_2=1.5$ for open-domain QA. 

For Poem generation, we design $l_{format}$ using various format information like number of characters in each short sentence, repetitive, rhythm, and tones. Despite it doesn't work when applied to prompting baseline directly, it cooperates well with \model . 

\subsection{Human Evaluation Details}
\begin{table*}[ht]

    \centering
    \caption{Performance and Deviation for open-domain long-form Chinese QA under Human Evaluation. }
    \label{tab:qa-eval-deviation}
    \vspace{-0.1in}
    \begin{threeparttable}
    \begin{tabular}{lccccc}
    \toprule
         \multirow{2}{*}{Method} & Fluency & Informativeness & Relevance & Overall \\
         & (1-5) & (1-5) & (1-5) & (1-10) \\
    \midrule
       CPM &2.66$\pm$0.19&2.47$\pm$0.19&2.36$\pm$0.20&4.32$\pm$0.37\\
        Prompting Baseline &3.44$\pm$0.19&3.25$\pm$0.20&3.21$\pm$0.22&5.97$\pm$0.42 \\
        Inverse Prompting &\textbf{3.61$\pm$0.17}&\textbf{3.43$\pm$0.19}&\textbf{3.59$\pm$0.20}&\textbf{6.51$\pm$0.38} \\
    \cmidrule{1-5}
        Human Answers &3.80$\pm$0.18&3.61$\pm$0.19&3.67$\pm$0.21&6.85$\pm$0.39\\
    \bottomrule
    \end{tabular}
    \end{threeparttable}
\end{table*}

\begin{figure}
    \centering
   
    \begin{subfigure}
    \centering
    \includegraphics[width=0.45\textwidth]{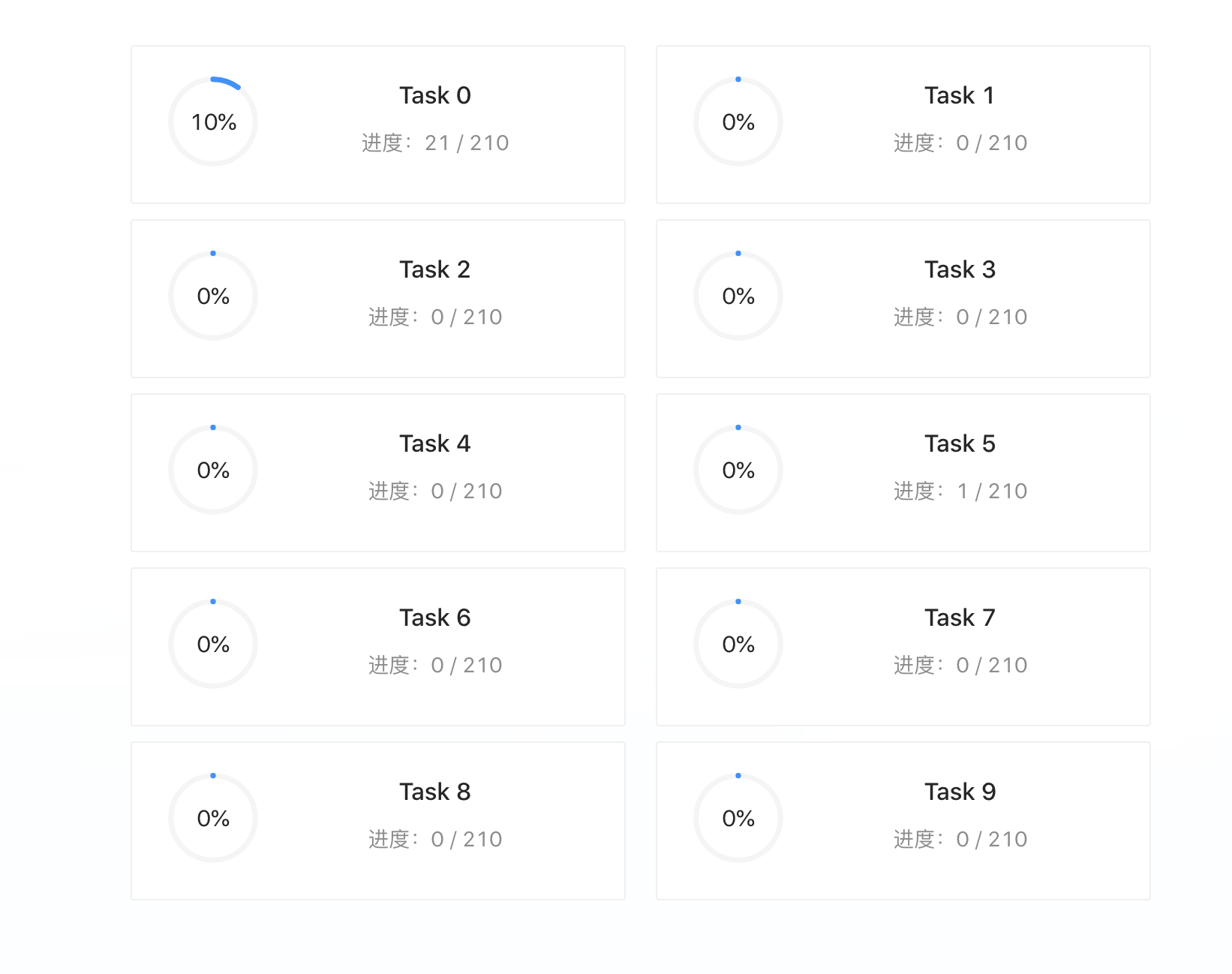}
    \end{subfigure}
    \begin{subfigure}
    \centering
    \includegraphics[width=0.45\textwidth]{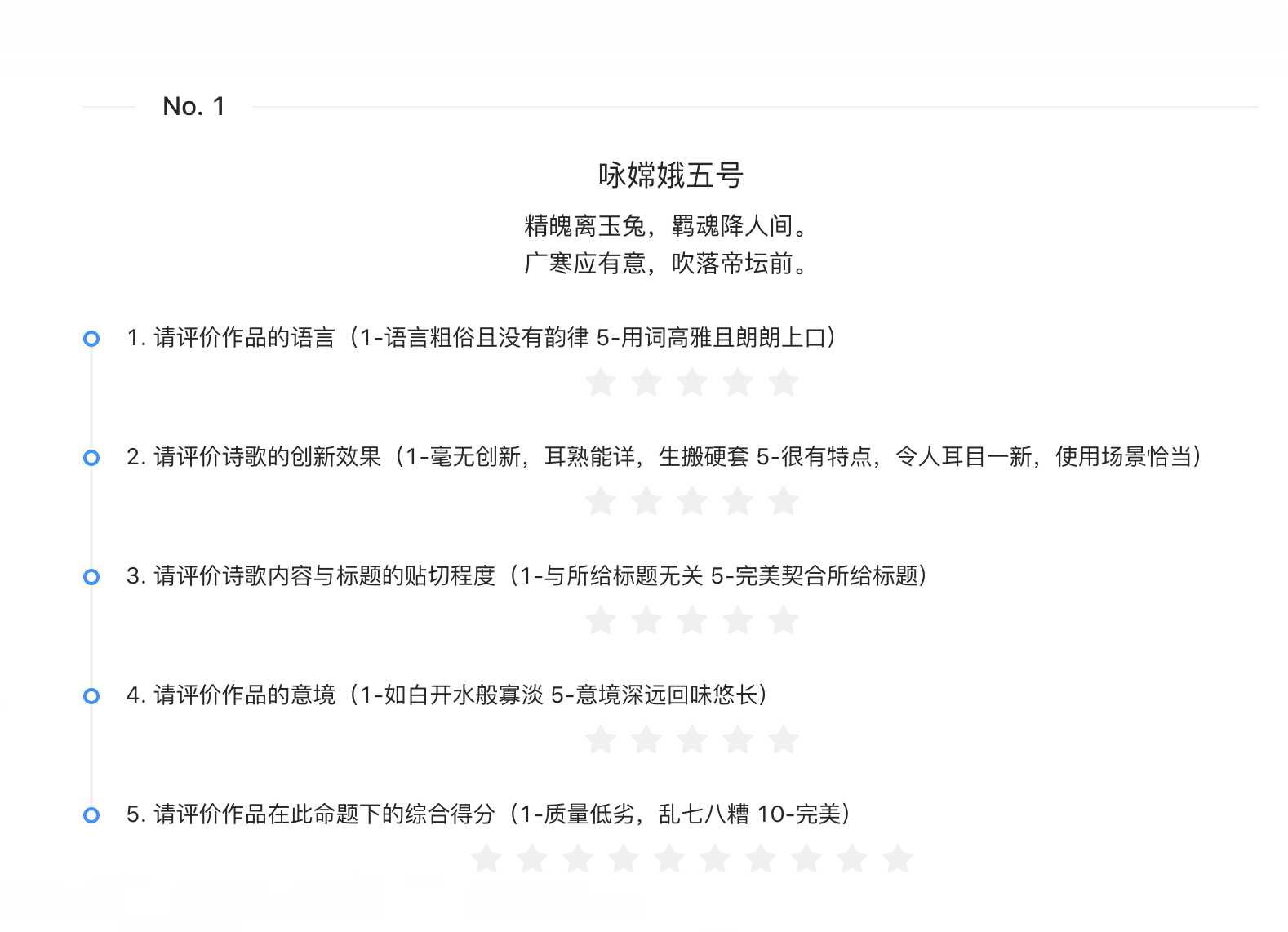}
    \end{subfigure}
    
    \caption{An illustration of our human evaluation platform. The whole task of evaluating 100 prompts is divided into 10 sub-tasks, and in each sub-task, the evaluator is required to score 4 contexts for 10 prompts in multiple aspects.}
    \label{fig:platform}
    
\end{figure}

\begin{table*}[ht]

    \centering
    \caption{Performance and Deviation for open-domain Traditional Chinese Poem Generation under human expert evaluation.  }
    \vspace{-0.1in}
    \label{tab:poem_eval_deviation}
    \begin{threeparttable}
    \begin{tabular}{lccccc}
    \toprule
         \multirow{2}{*}{Method} & Format & Innovation & Relevance & Aesthetics & Overall \\
         & (1-5) & (1-5) & (1-5) & (1-5) & (1-10) \\
    \midrule
      Jiuge &\textbf{3.60$\pm$0.25}&2.47$\pm$0.28&1.99$\pm$0.31&\textbf{3.12$\pm$0.31}&3.57$\pm$0.54\\
        Search Baseline &2.79$\pm$0.37&1.10$\pm$0.13&1.16$\pm$0.16&2.44$\pm$0.38 &1.35$\pm$0.27  \\
    \cmidrule{1-6}
        Inverse Prompting & 2.56$\pm$0.28&2.71$\pm$0.28&2.92$\pm$0.37&2.33$\pm$0.28&4.00$\pm$0.52  \\
        Inverse Prompting +ST & 2.42$\pm$0.29&\textbf{2.92$\pm$0.28}&\textbf{3.65$\pm$0.33}&2.18$\pm$0.28& \textbf{4.40$\pm$0.47} \\
    \bottomrule
    \end{tabular}
    \end{threeparttable}
\end{table*}
Our human evaluation is conducted on a platform. Figure~\ref{fig:platform} illustrates how the evaluation platform looks like. The whole task of evaluating 100 prompts is divided into 10 sub-tasks, and in each sub-task, the evaluator is required to score 4 contexts for 10 prompts in multiple aspects, like an online questionnaire.

The evaluation does not necessarily need to be finished at once. People can login and logout, change their answers for already completed problems, or continue evaluation from their current positions freely in one week's time. They only need to ensure that all evaluation questions have been answered before the deadline, with the ratings being consistent. 

Valid evaluators for open-domain QA are paid 150 RMB yuan each (about $\$25$), while each TCP evaluator receives 300 RMB yuan (about $\$50$), as evaluation for traditional poems requires more expert reviewers. The payment is not high but due to the flexible time arrangement for online and interesting content, the task still attracted a lot of participants.

We recruit 11 experts for TCP evaluation, 10 of them finished and all of those finished provide valid evaluations, we recruit 45 people for open-domain QA, 32 of them finish their experiments and 30 of them provide consistent evaluations.

For generating baseline texts, for QA, we generate the prompting baseline using the base text generation code under the prompt format of "Question:$\$\textbf{Question}$ Description:$\$\textbf{Description}$ Answer:", for CPM we apply the same prompt format and use its recommended text generation code. 

For poem generation using Jiuge, we write code that can automatically make online queries to  its public generation website \url{http://jiuge.thunlp.org/} and get generated poems. Jiuge has a lot of format modes and we choose four most general modes without additional restrictions  "5-Jueju","7-Jueju","5-Lvshi","7-Lvshi". For each title Jiuge generates one best poem for each mode. However, it offers no hint about which poem it considers the best so we randomly choose one from the 4 generated for human evaluation. 

\subsection{Deviation for Human Evaluators and p-values}
\label{app:p-value}
Table \ref{tab:qa-eval-deviation},\ref{tab:poem_eval_deviation} displays the deviation of the scorings for human evaluators. 

The deviation is calculated in a per-capita basis that we first average the scorings for each method on for every evaluator, then we compute the deviation based on the average scores of each human evaluators. 

As can be seen, evaluators agree more on the quality for Chinese QA, while less agree on the qualities for poems. 

With those standard deviations, assuming evaluators are independent with each other, we can calculate p-score. For poems we have $N=10$, the p-value for Jiuge $\geq$ Inverse Prompting is 0.0544 while the p-value for Jiuge $\geq$ Inverse Prompting+self-training is 0.0009, suggesting that under $p<.05$ we cannot fully reject the hypothesis that Jiuge is not worse to Inverse Prompting. However, Inverse Prompting with self-training is statistically better than Jiuge.

For QA, with $N=30$ the p-value for Prompting Baseline $\geq$ Inverse Prompting is $<.00001$, while the p-value for Inverse Prompting $\geq$ Human is 0.0006. So \model is statistically better than the prompting baseline but is still worse than human.

\subsection{Online Demo Platforms}
We further developed the poem generation and add some other formats of poems, including heading, which pre-defines the first word of each short sentence before poem generation, and SongCi, which is another form of traditional Chinese context that involves much higher format standard. All of these downstream tasks are based on the inverse prompting+self training protocol , with tiny format adjustments for each downstream task. 

 We display these applications on our demo Wudao Poetry \footnote{\url{https://pretrain.aminer.cn/apps/poetry.html}}. Users can also submit their customized titles and generate poems of their own. There is also a QA demo named Wudao QA  \footnote{\url{https://pretrain.aminer.cn/os/qa}} where users can submit their own questions and descriptions to get an AI answer.
 
  \begin{figure}
     \centering
     \includegraphics[width=0.35\textwidth]{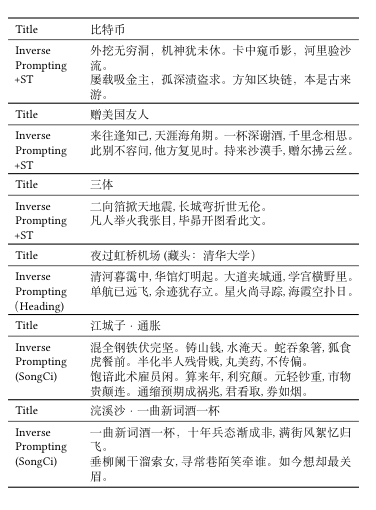}
     \caption{Selected examples of \model poetry, poems with heading, and SongCi.}
     \label{fig:more}
 \end{figure}
 
 Figure \ref{fig:more} displays some of the generated poems for these downstream tasks on the platform. More cases can be found on the platform, or generated according to users' submissions.

\clearpage
\newpage
\bibliographystyle{ACM-Reference-Format}
\bibliography{reference.bib}

\end{document}